\documentclass{article}

\usepackage[final]{corl_2019_preprint} 

\usepackage{graphics} 
\usepackage{amsmath, bm} 
\usepackage{amssymb}  
\usepackage{graphicx}
\usepackage{hyperref}
\usepackage{subcaption}
\usepackage{graphicx}
\usepackage{mathtools}
\usepackage{dsfont}
\usepackage{float}
\usepackage{makecell}
\usepackage{authblk}
\usepackage{algorithm}
\usepackage{algcompatible}
\usepackage{graphicx}
\usepackage{multirow} 
\usepackage{wrapfig}
\usepackage[table]{xcolor}
\renewcommand{\baselinestretch}{1.0}

\makeatletter
\renewcommand\AB@affilsepx{\hspace{1in} \protect\Affilfont}
\makeatother

\newcommand\blfootnote[1]{%
  \begingroup
  \renewcommand\thefootnote{}\footnote{#1}%
  \addtocounter{footnote}{-1}%
  \endgroup
}

\usepackage{subcaption}
\DeclareMathOperator{\E}{\mathbb{E}}

\title{Deep Reinforcement Learning for Industrial Insertion Tasks with Visual Inputs and Natural Rewards
}

\author{Gerrit Schoettler$^{*1}$, Ashvin Nair$^{*2}$, Jianlan Luo$^{2}$, Shikhar Bahl$^{2}$,\\ Juan Aparicio Ojea$^{1}$, Eugen Solowjow$^{1}$, Sergey Levine$^2$}

\begin{document}


\vspace{-10pt}
\maketitle
\blfootnote{$^*$ First two authors contributed equally, $^1$ Siemens Corporation, $^2$ University of California, Berkeley.}
\begin{abstract}
Connector insertion and many other tasks commonly found in modern manufacturing settings involve complex contact dynamics and friction.
Since it is difficult to capture related physical effects with first-order modeling, traditional control methods often result in brittle and inaccurate controllers, which have to be manually tuned.
Reinforcement learning (RL) methods have been demonstrated to be capable of learning controllers in such environments from autonomous interaction with the environment, but running RL algorithms in the real world poses sample efficiency and safety challenges.
Moreover, in practical real-world settings we cannot assume access to perfect state information or dense reward signals.
In this paper, we consider a variety of difficult industrial insertion tasks with visual inputs and different natural reward specifications, namely sparse rewards and goal images.
We show that methods that combine RL with prior information, such as classical controllers or demonstrations, can solve these tasks from a reasonable amount of real-world interaction.

\end{abstract}



\section{Introduction}\label{sec:introduction}

Many industrial tasks on the edge of automation require a degree of adaptability that is difficult to achieve with conventional robotic automation techniques.
While standard control methods, such as PID controllers, are heavily employed to automate many tasks in the context of positioning, tasks that require significant adaptability or tight visual perception-control loops are often beyond the capabilities of such methods, and therefore are typically performed manually.
Standard control methods can struggle in presence of complex dynamical phenomena that are hard to model analytically, such as complex contacts.
Reinforcement learning (RL) offers a different solution, relying on trial and error learning instead of accurate modeling to construct an effective controller.
RL with expressive function approximation, i.e. deep RL, has further shown to automatically handle high dimensional inputs such as images \cite{mnih2013atari}.

However, deep RL has thus far not seen wide adoption in the automation community due to several practical obstacles. 
Sample efficiency is one obstacle: tasks must be completed without excessive interaction time or wear and tear on the robot. Progress in recent years on developing better RL algorithms has led to significantly better sample efficiency, even in dynamically complicated tasks \cite{haarnoja2018sac, hessel2018rainbow},
but remains a challenge for deploying RL in real-world robotics contexts.
Another major, often underappreciated, obstacle is goal specification: while prior work in RL assumes a reward signal to optimize,
it is often carefully shaped to allow the system to learn \cite{ng1999rewardshaping, popov17stacking, daniel2014activereward}. 
Obtaining such dense reward signals can be a significant challenge, as one must additionally build a perception system that allows computing dense rewards on state representations. Shaping a reward function so that an agent can learn from it is also a manual process that requires considerable manual effort. An ideal RL system would learn from rewards that are natural and easy to specify.
How can we enable robots to autonomously perform complex tasks without significant engineering effort to design perception and reward systems?

\begin{figure}[t]
    \centering
    \begin{subfigure}{0.14\linewidth}
        \center
        USB \vspace{1.1cm}
        
        D-Sub \vspace{1.1cm}
        
        Model-E
    \end{subfigure}
    \begin{subfigure}{0.14\linewidth}
        \center
        \includegraphics[height=1.17cm]{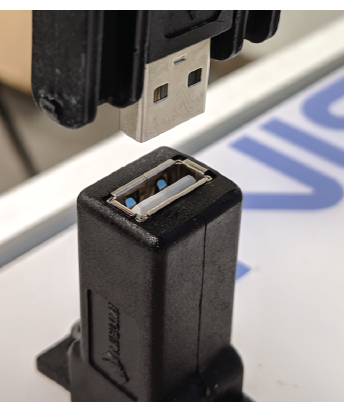} \vspace{0.2cm}
        
        \includegraphics[height=1.17cm]{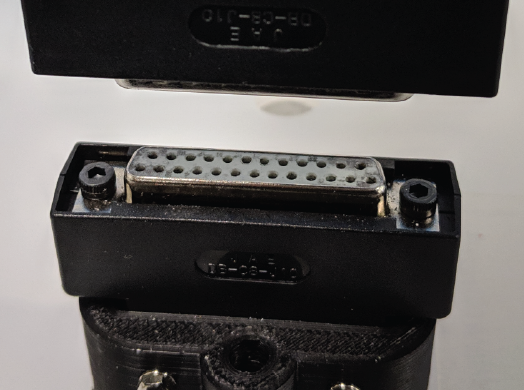} \vspace{0.2cm}
        
        \includegraphics[height=1.17cm]{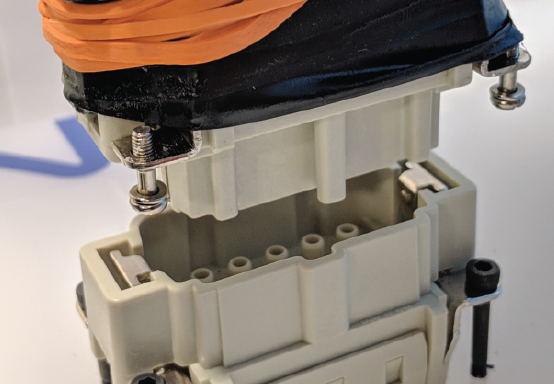} \vspace{0.05cm}
    \end{subfigure}
    \begin{subfigure}{0.69\linewidth}
        \begin{subfigure}[b]{0.69\linewidth}
            \includegraphics[trim={1.7cm 0 0 0},clip,width=0.99\linewidth]{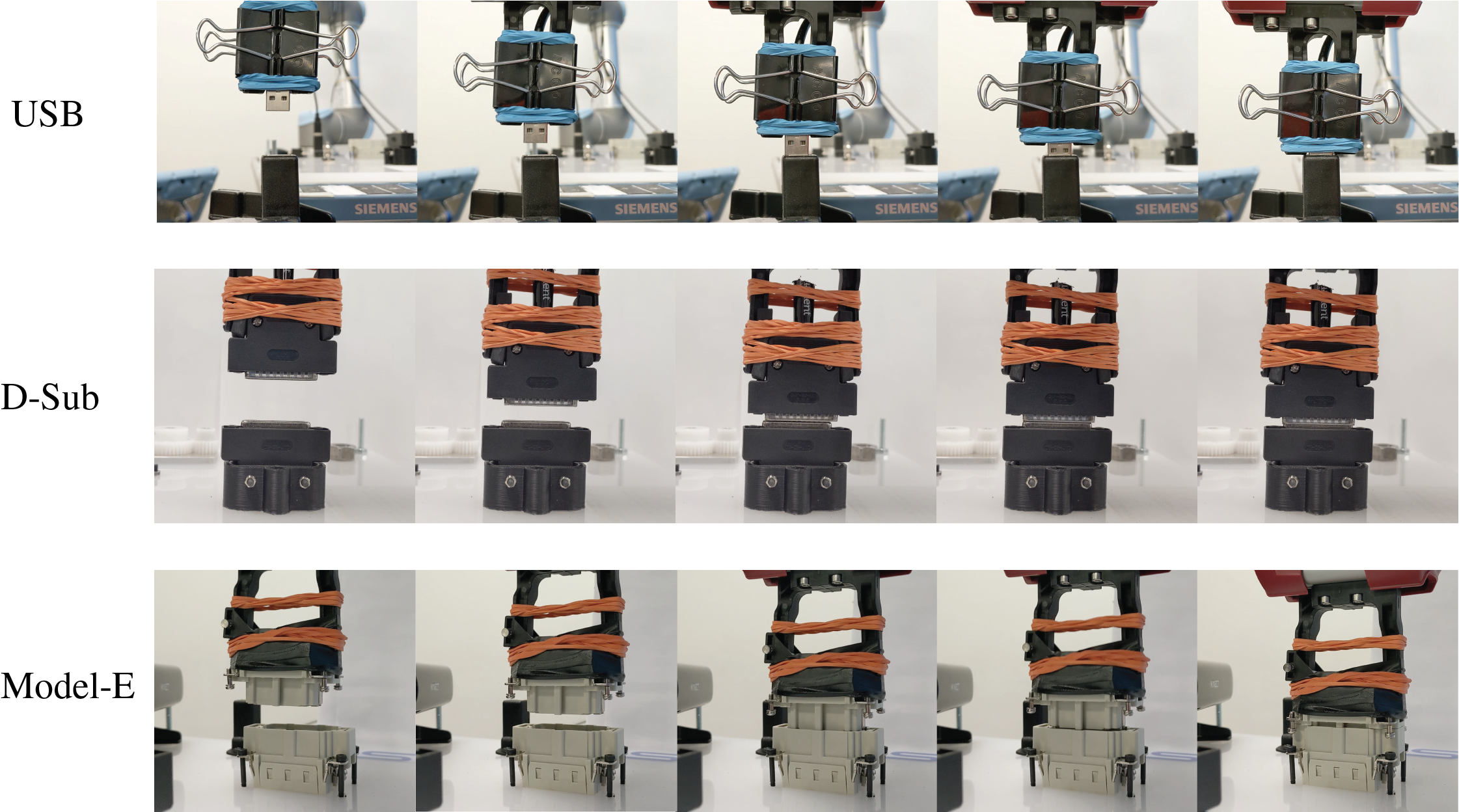} 
            \centering
        \end{subfigure}\hspace{2pt}
        \begin{subfigure}[b]{0.29\linewidth}
            \includegraphics[trim={0 0 9cm 0},clip,width=0.99\linewidth]{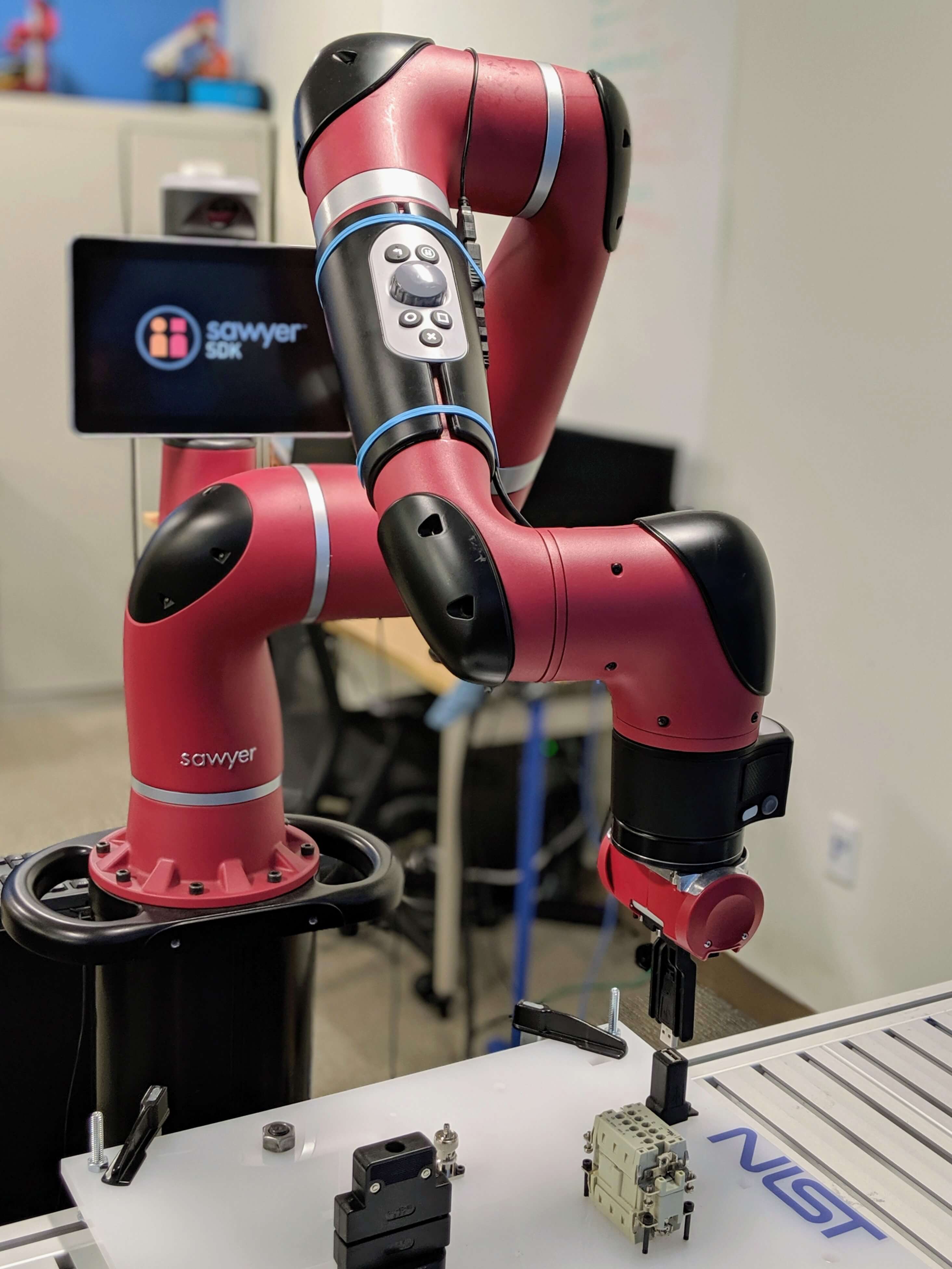}
            \centering
        \end{subfigure}\hfill
    \end{subfigure}

\caption{We train policies directly in the real world to solve connector insertion tasks from raw pixel input and without access to ground-truth state information for reward functions. Left: top-down views of the connectors. Middle: a rollout from a learned policy that successfully completes the insertion task for each connector is shown. Right: a full view of the robot setup. Videos of the results are available at \href{https://industrial-insertion-rl.github.io/}{industrial-insertion-rl.github.io} }
    \label{fig:usb_photo_demo2}
\end{figure}

We first consider an end-to-end approach that learns a policy from images, where the images serve as both the state representation and the goal specification. Using goal images is not fully general, but can successfully represent tasks when the task is to reach a final desired state \cite{nair2018rig}.
Specifying goals via goal images is convenient, and makes it possible to specify goals with minimal manual effort. Using images as the state representation also allows a robot to learn behaviors that utilize direct visual feedback, which provides some robustness to sensor and actuator noise.

Secondly, we consider learning from simple and sparse reward signals. Sparse rewards can often be obtained conveniently, for instance from human-provided labels or simple instrumentation. In many electronic assembly tasks, which we consider here, we can directly detect whether the electronics are functional, and use that signal as a reward. Learning from sparse rewards poses a challenge, as exploration with sparse reward signals is difficult, but by using sufficient prior information about the task, one can overcome this challenge. To handle this challenge, we extend the residual RL approach~\cite{johannink18residualrl, silver18residualpolicylearning}, which learns a parametric policy on top of a fixed, hand-specified controller, to the setting of vision-based manipulation.

    

In our experiments, we show that we can successfully complete real-world tight tolerance assembly tasks, such as inserting USB connectors, using RL from images with reward signals that are convenient for users to specify.
We can learn from only a sparse reward based on the electrical connection for a USB adapter plug, and we demonstrate learning insertion skills with rewards based only on goal images.
These reward signals require no extra engineering and are easy to specify for many tasks. 
Beyond showing the feasibility of RL for solving these tasks, we evaluate multiple RL algorithms across three tasks and study their robustness to imprecise positioning and noise.

\section{Related Work}\label{sec:relatedwork}

Learning has been applied previously in a variety of robotics contexts. Different forms of learning have enabled autonomous driving \cite{pomerleau1989alvinn}, biped locomotion \cite{nakanishi2004bipedlfd}, block stacking \cite{deisenroth2011stacking}, grasping \cite{pinto2015supersizing}, and navigation \cite{giusti15trails, pathak2018zeroshot}. Among these methods, many involve reinforcement learning, where an agent learns to perform a task by maximizing a reward signal. Reinforcement learning algorithms have been developed and applied to teach robots to perform tasks such as balancing a robot \cite{deisenroth2011pilco}, playing ping-pong \cite{peters2010reps} and baseball \cite{peters2008baseball}.
The use of large function approximators, such as neural networks, in RL has further broadened the generality of RL \cite{mnih2013atari}. Such techniques, called ``deep'' RL, have further allowed robots to be trained directly in the real world to perform fine-grained manipulation tasks from vision \cite{levine2016gps}, open doors \cite{gu2016naf}, play hockey \cite{chebotar2017pilqr}, stack Lego blocks \cite{zhang2019solar}, use dexterous hands \cite{zhu2019hands}, and grasp objects \cite{kalashnikov2018qtopt}. In this work we further explore solving real-world robotics tasks using RL.

Many RL algorithms introduce prior information about the specific task to be solved. One common method is reward shaping \cite{ng1999rewardshaping}, but reward shaping can become arbitrarily difficult as the complexity of the task increases. Other methods incorporate a trajectory planner \cite{thomas2018cad} but for complex assembly tasks, trajectory planners require a host of information about objects and geometries which can be difficult to provide.

Another body of work on incorporating prior information studies using  demonstrations either to initialize a policy \cite{peters2008baseball, kober2008mp}, infer reward functions using inverse reinforcement learning \cite{finn16guidedcostlearning, ziebart2008maxent} or to improve the policy throughout the learning procedure \cite{hester17dqfd, nair2018demonstrations, rajeswaran2018dextrous}. These methods require multiple demonstrations, which can be difficult to collect, especially for assembly tasks, although learning a reward function by classifying goal states \cite{singh2019raq} may partially alleviate this issue. More recently, manually specifying a policy and learning the residual task has been proposed \cite{johannink18residualrl, silver18residualpolicylearning}. In this work we evaluate both residual RL and combining RL with learning from demonstrations.

Previous work has also tackled high precision assembly tasks, especially insertion-type tasks. One line of work focuses on obtaining high dimensional observations, including geometry, forces, joint positions and velocities \cite{li2014usbgelsight, tamar2017hindsightplan, inoue2017deeprlassembly, luo19variableimpedance}, but this information is not easily procured, increasing complexity of the experiments and the supervision required. Other work relies on external trajectory planning or very high precision control \cite{inoue2017deeprlassembly, tamar2017hindsightplan}, but this can be brittle to error in other components of the system, such as perception. We show how our method not only solves insertion tasks with much less information about the environment, but also does so under noisy conditions. 
\section{Electric Connector Plug Insertion Tasks}

\begin{wrapfigure}{r}{0.5\textwidth}
  \begin{center}
        \includegraphics[width=0.88\linewidth]{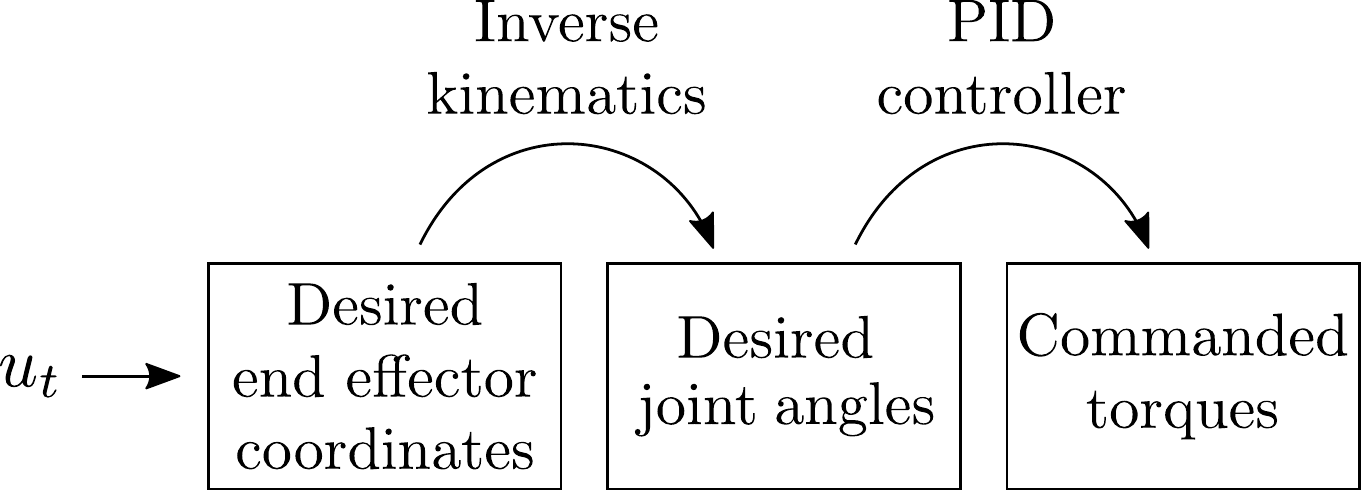}
    \caption{Illustration of the robot's cascade control scheme. The actions  $u_t$ are computed at a frequency of up to $10\,\mathrm{Hz}$,  desired joint angles are obtained by inverse kinematics, and a joint-space impedance controller with anti-windup PID control commands actuator torques at $1000\,\mathrm{Hz}$.}
   \label{fig:control_pipeline}
  \end{center}
\end{wrapfigure}

In this work, we empirically evaluate learning methods on a set of electric connector assembly tasks, pictured in Fig.~\ref{fig:usb_photo_demo2}.
Connector plug insertions are difficult for two reasons. 
First, the robot must be very precise in lining up the plug with its socket. 
As we show in our experiments, errors as small as~$\pm 1$\,mm can lead to consistent failure. 
Second, there is significant friction when the connector plug touches the socket, and the robot must learn to apply sufficient force in order to insert the plug. Image sequences of successful insertions are shown in Fig.~\ref{fig:usb_photo_demo2}, where it is also possible to see details of the gripper setup that we used to ensure a failure free, fully automated training process.
In our experiments, we use a 7 degrees of freedom Sawyer robot with end-effector control, meaning that the action signal $u_t$ can be interpreted as the relative end-effector movement in Cartesian coordinates. The robot's underlying internal control pipeline is illustrated in Figure~\ref{fig:control_pipeline}.

To comprehensively evaluate connector assembly tasks, we experiment on a variety of connectors. Each connector offers a different challenge in terms of required precision and force to overcome friction. 
We chose to benchmark the controllers performance on the insertion of a USB connector, a U-Sub connector, and a waterproof Model-E connector manufactured by MISUMI.
All the explored use cases were part of the IROS 2017 Robotic Grasping and Manipulation Competition \cite{roboticgrasping2017iros}, included as part of a task board developed by NIST to benchmark the performance of assembly robots.

\subsection{Adapters}
In the following we describe the used adapters, USB, D-Sub, and Model-E. The observed difficulty of the insertion increases in that order.

\textbf{USB.} The USB connector is a ubiquitous, widely-used connector and offers a challenging insertion task. Because the adapter becomes smoother and therefore easier to insert over time due to wear and tear, we periodically replace the adapter. Of the three tested adapters, the USB adapter is the easiest.

\textbf{D-sub.} Inserting this adapter requires aligning several pins correctly, and is therefore more sensitive than inserting the USB adapter. It also requires more downward force due to a tighter fit.

\textbf{Model-E.} This adapter is the most difficult of the three tested connectors as it contains several edges and grooves to align and requires significant downward force to successfully insert the part.

\subsection{Experimental Settings}
We consider three settings in our experiments in order to evaluate how plausible it is to solve these tasks with more convenient state representations and reward functions and to evaluate the performance of different algorithms changes as the setting is modified.

\begin{wrapfigure}{r}{0.5\textwidth}
\centering
\resizebox{.95\linewidth}{!}{
    \begin{subfigure}[b]{0.24\linewidth}
        \includegraphics[width=0.99\linewidth]{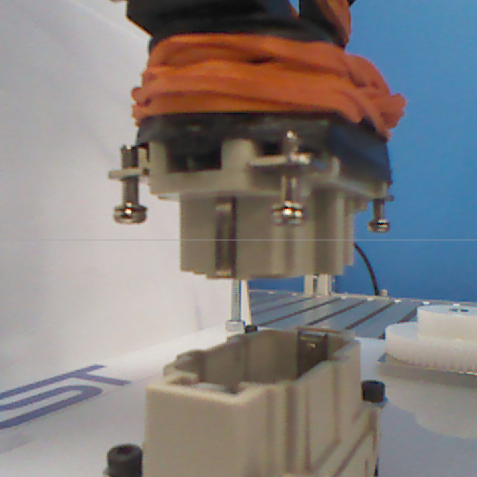}\\
        \centering
    \end{subfigure}
    \begin{subfigure}[b]{0.24\linewidth}
        \includegraphics[width=0.99\linewidth]{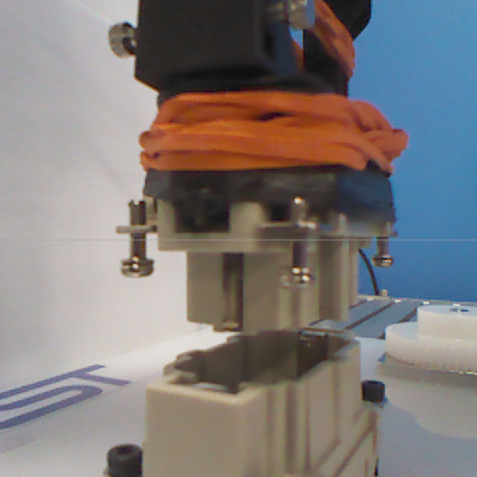}\\
        \centering
    \end{subfigure}  
    \begin{subfigure}[b]{0.24\linewidth}
        \includegraphics[width=0.99\linewidth]{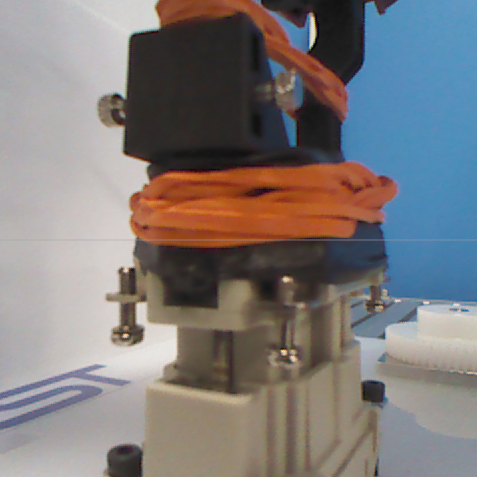}\\
        \centering
    \end{subfigure} 
    \begin{subfigure}[b]{0.24\linewidth}
        \includegraphics[width=0.99\linewidth]{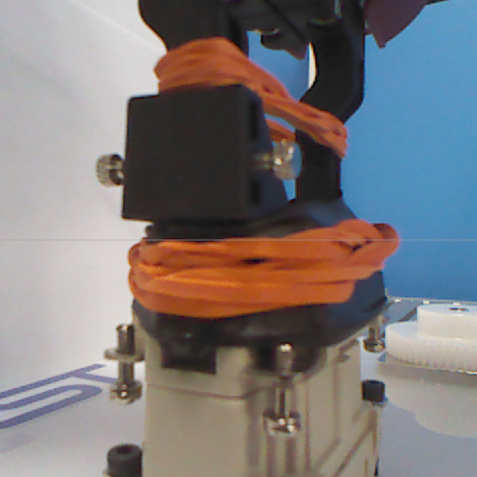}\\
        \centering
    \end{subfigure}
    }
    \\
    \vspace{-6pt}
    \resizebox{.95\linewidth}{!}{
    \begin{subfigure}[b]{0.24\linewidth}
        \includegraphics[width=0.99\linewidth]{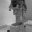}\\
        \centering
    \end{subfigure} 
    \begin{subfigure}[b]{0.24\linewidth}
        \includegraphics[width=0.99\linewidth]{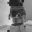}\\
        \centering
    \end{subfigure}     
    \begin{subfigure}[b]{0.24\linewidth}
        \includegraphics[width=0.99\linewidth]{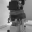}\\
        \centering
    \end{subfigure} 
    \begin{subfigure}[b]{0.24\linewidth}
        \includegraphics[width=0.99\linewidth]{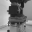}\\
        \centering
    \end{subfigure}   
     }
    \caption{Successful insertion on the Model-E connector. The image-based RL algorithms receives only receives the $32\times32$ grayscale image as the observation.
    }
    \label{fig:vision_insertion_sequence}
\end{wrapfigure}

\textbf{3.2.1 Visual. } In this experiment, we evaluate whether the RL algorithms can learn to perform the connector assembly tasks from vision without having access to state information. The state provided to the learned policy is a $32 \times 32$ grayscale image, such as shown in Fig.~\ref{fig:vision_insertion_sequence}.
For goal specification, we use a goal image, avoiding the need for state information to compute rewards. The reward is the pixelwise L1 distance to the given goal image. Being able to learn from such a setup is compelling as it does not require any extra state estimation and many tasks can be specified easily by a goal image.

\textbf{3.2.2. Sparse. } In this experiment, the reward is obtained by directly measuring whether the connection is alive and transmitting:
\begin{align}
 r  = 
  \begin{cases}
    1, & \text{if insertion signal detected } \\
    0, & \text{else. } \\
  \end{cases}
\end{align}
This is the exact true reward for the task of connecting a cable, and can be naturally obtained in many manufacturing systems. As state, the robot is given the Cartesian coordinates of the end-effector $x_t$ and the vertical force $f_z$ that is acting on the end-effector. We could only automatically detect the USB connection, so we only include the USB adapter for the sparse experiments. 

\textbf{3.2.3. Dense. } In this experiment, the robot receives a manually shaped reward based on the distance to the target location $ x^{*}$. We use the reward function
\begin{equation}
r_t =  - \alpha \cdot {\Vert x_t - x^{*} \Vert}_1 -  \frac{\beta}{\left({\Vert x_t - x^{*} \Vert}_2 + \varepsilon\right)} - \varphi \cdot f_z,
\label{eq:shaped_reward_function}
\end{equation}
where $0 < \varepsilon \ll 1$. 
The hyperparameters are set to ${\alpha = 100}$, ${\beta = 0.002}$, and ${\varphi = 0.1}$. When an insertion is indicated through a distance measurement, the sign of the force term flips, so that $\varphi = -0.1$ when the connector is inserted. This rewards the agent for pressing down after an insertion and showed to improve the learning process.
The force measurements are calibrated before each rollout to account for measurement bias and to decouple the measurements from the robot pose. 

\section{Methods}\label{sec:method}

To solve the connector insertion tasks, we consider and evaluate a variety of RL algorithms.

\subsection{Preliminaries}\label{sec:background}
In a Markov decision process (MDP), an agent at every time step is at state $s_t \in \mathcal{S}$, takes actions $u_t \in \mathcal{U}$, receives a reward $r_t \in \mathbb{R}$, and the state evolves according to environment transition dynamics $p(s_{t+1}|s_t, u_t)$. The goal of reinforcement learning is to choose actions $u_t \sim \pi(u_t|s_t)$ to maximize the expected returns $\E[\sum_{t=0}^H \gamma^t r_t]$ where $H$ is the horizon and $\gamma$ is a discount factor. The policy $\pi(u_t|s_t)$ is often chosen to be an expressive parametric function approximator, such as a neural network, as we use in this work.

\subsection{Efficient Off-Policy Reinforcement Learning}
One class of RL methods additionally estimates the expected discounted return after taking action $u$ from state $s$, the {Q-value} $Q(s, u)$. Q-values can be recursively defined with the Bellman equation:
\begin{align}
    Q(s_t, u_t) = \E_{s_{t+1}}[r_t + \gamma \max_{u_{t+1}} Q(s_{t+1}, u_{t+1})]
\end{align}
and learned from off-policy transitions $(s_t, u_t, r_t, s_{t+1})$. Because we are interested in sample-efficient real-world learning, we use such RL algorithms that can take advantage of off-policy data.

For control with continuous actions, computing the required maximum in the Bellman equation is difficult. Continuous control algorithms such as deep deterministic policy gradients (DDPG) \cite{lillicrap2015continuous} additionally learn a policy $\pi_\theta(u_t|s_t)$ to approximately choose the maximizing action. In this paper we specifically consider two related reinforcement learning algorithms that lend themselves well to real-world learning as they are sample efficient, stable, and require little hyperparameter tuning.

\textbf{Twin Delayed Deep Deterministic Policy Gradients (TD3).} 
Like DDPG, TD3 optimizes a deterministic policy \cite{fujimoto2018td3} but uses two Q-function approximators to reduce value overestimation \cite{vanhasselt2016doubledqn} and delayed policy updates to stabilize training.

\textbf{Soft Actor Critic (SAC).}
SAC is an off-policy value-based reinforcement learning method based on the maximum entropy reinforcement learning framework with a stochastic policy \cite{haarnoja2018sac}.

We used the implementation of these RL algorithms publicly available at \href{https://github.com/vitchyr/rlkit}{\texttt{rlkit}} \cite{pong2018tdm}.

\subsection{Residual Reinforcement Learning}

Instead of randomly exploring from scratch, we can inject prior information into an RL algorithm in order to speed up the training process, as well as to minimize unsafe exploration behavior. In residual RL, actions $u_t$ are chosen by additively combining a fixed policy $\pi_\text{H}(s_t)$ with a parametric policy $\pi_\theta(u_t|s_t)$:
\begin{equation}\label{eq:ctrl_seq}
    u_t = \pi_\text{H}(s_t) + \pi_\theta(s_t).
\end{equation}

The parameters $\theta$ can be learned using any RL algorithm. In this work, we evaluate both SAC and TD3, explained in the previous section. The residual RL implementation that we use in our experiments is summarized in Algorithm~\ref{alg:residualrl}.

\begin{wrapfigure}{r}{0.55\textwidth}
    \begin{minipage}[t]{0.99\linewidth}
        \begin{algorithm}[H]
           	\caption{Residual reinforcement learning}
           	\label{alg:residualrl}
           	\begin{algorithmic}[1]
            \REQUIRE policy $\pi_\theta$, hand-engineered controller $\pi_\text{H}$.
            \FOR{$n=0,...,N-1$ episodes}
                \STATE Sample initial state $s_0 \sim E$.
                \FOR{$t=0,...,H -1$ steps}
                    \STATE Get policy action $u_t \sim \pi_\theta(u_t|s_t)$.
                    \STATE Get action to execute $u'_t = u_t + \pi_\text{H}(s_t)$.
                    \STATE Get next state $s_{t+1} \sim p(\cdot \mid s_t, u'_t)$.
                    \STATE Store $(s_t, u_t, s_{t+1})$ into replay buffer $\mathcal R$.
                    \STATE Sample set of transitions $(s, u, s') \sim \mathcal R$.
                    \STATE Optimize $\theta$ using RL with transitions.
                \ENDFOR
            \ENDFOR
           	\end{algorithmic}
        \end{algorithm}
        \vspace{1cm}
    \end{minipage}
\end{wrapfigure}

A simple P-controller serves as the hand-designed controller $\pi_\text{H}$ of our experiments.  The P-controller operates in Cartesian space and calculates the current control action by 
\begin{equation}
\pi_\text{H}(s_t) = - k_p\cdot (x_t - x^{*}), 
\end{equation}
where $x^{*}$ denotes the commanded goal location. As control gains we use $k_p = [\,1, \,1,\, 0.3\,]$. This P-controller quickly centers the end-effector above the goal position and reaches the goal after about 10 time steps from the reset position, which is located 5cm above the goal.

\begin{figure}[btp]
    \centering
    \resizebox{1.02\linewidth}{!}{
    \begin{subfigure}[b]{0.32\linewidth}
        \includegraphics[width=0.99\linewidth]{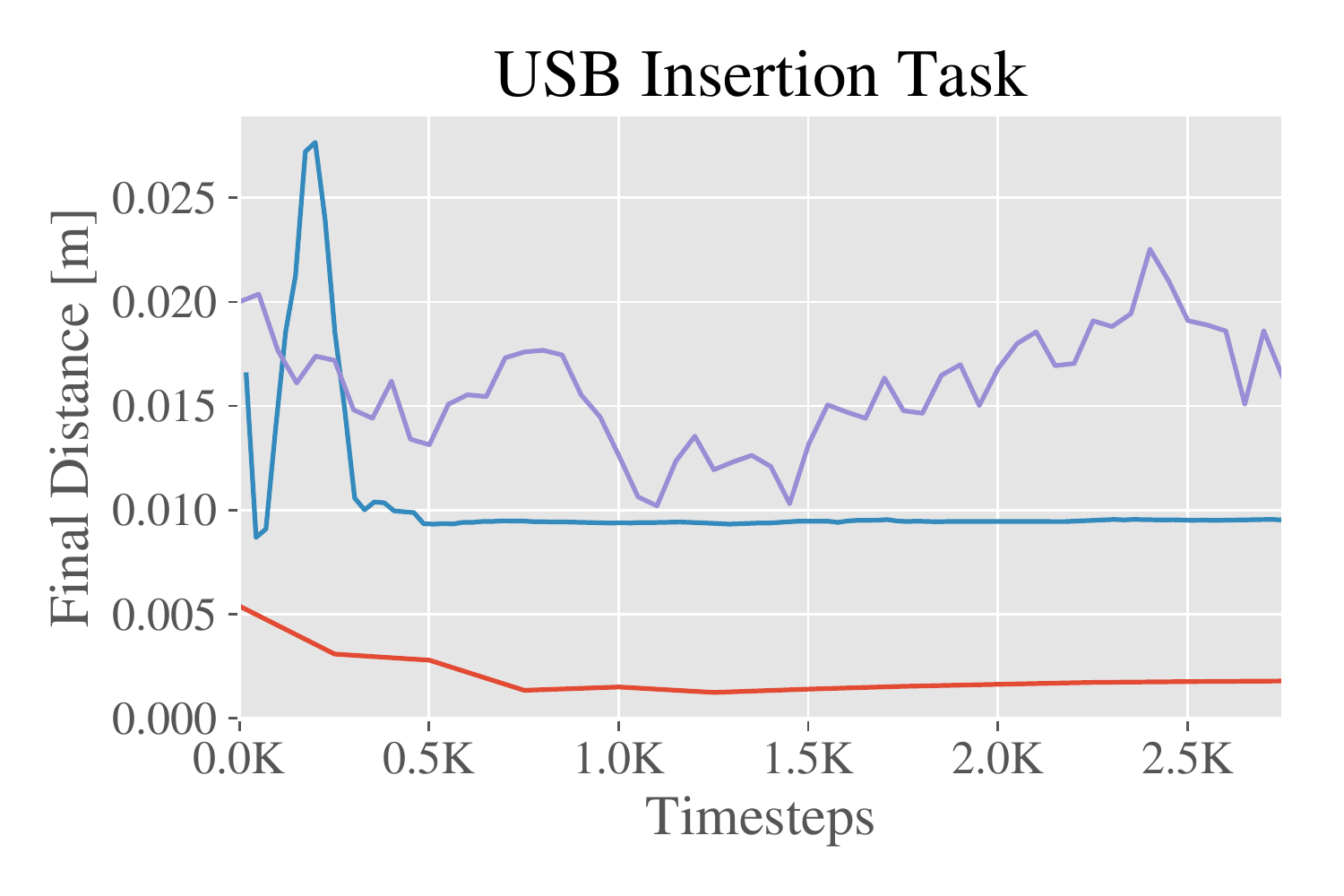}
    \end{subfigure}
    \begin{subfigure}[b]{0.32\linewidth}
        \includegraphics[width=0.99\linewidth]{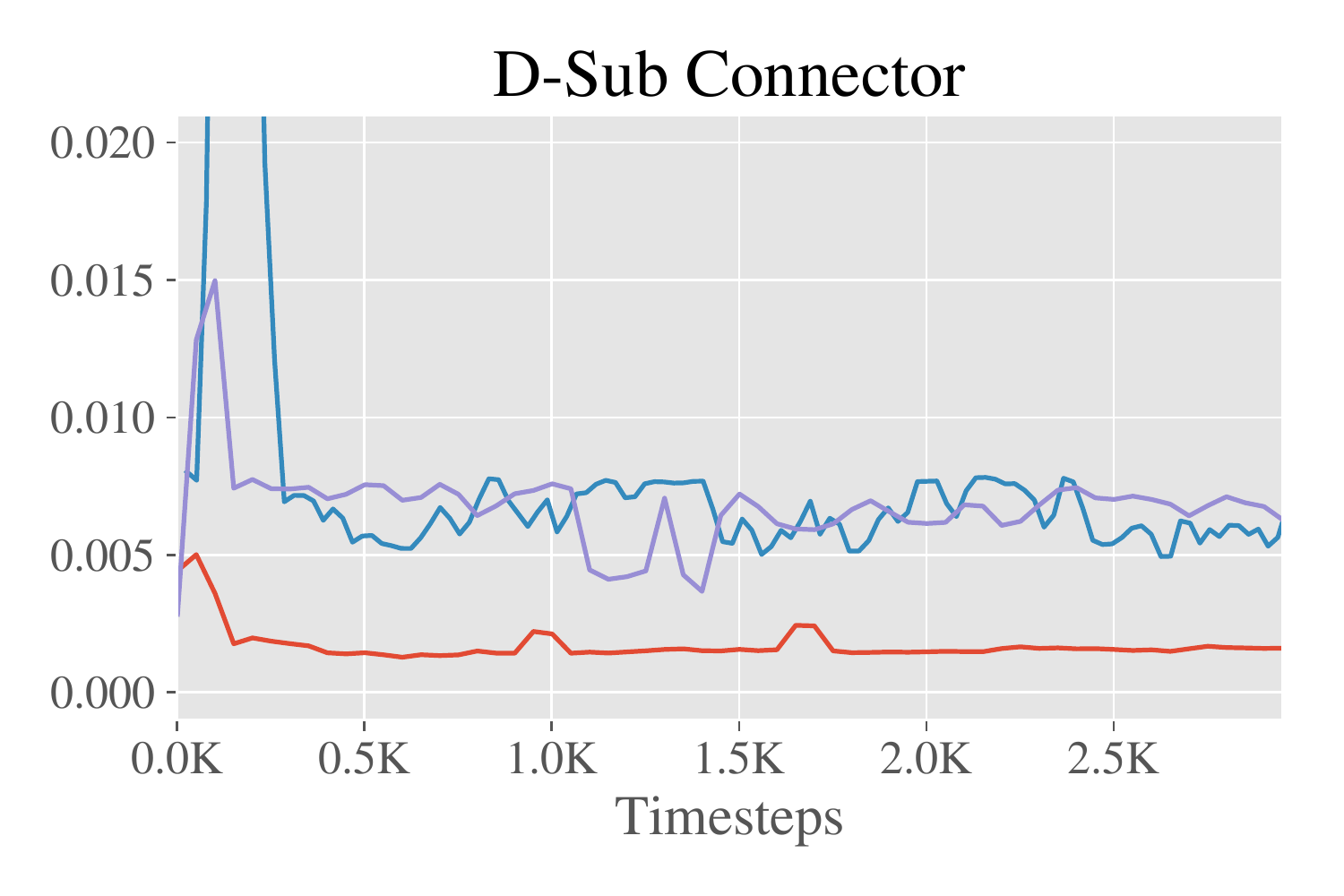}
    \end{subfigure}
    \begin{subfigure}[b]{0.32\linewidth}
        \includegraphics[width=0.99\linewidth]{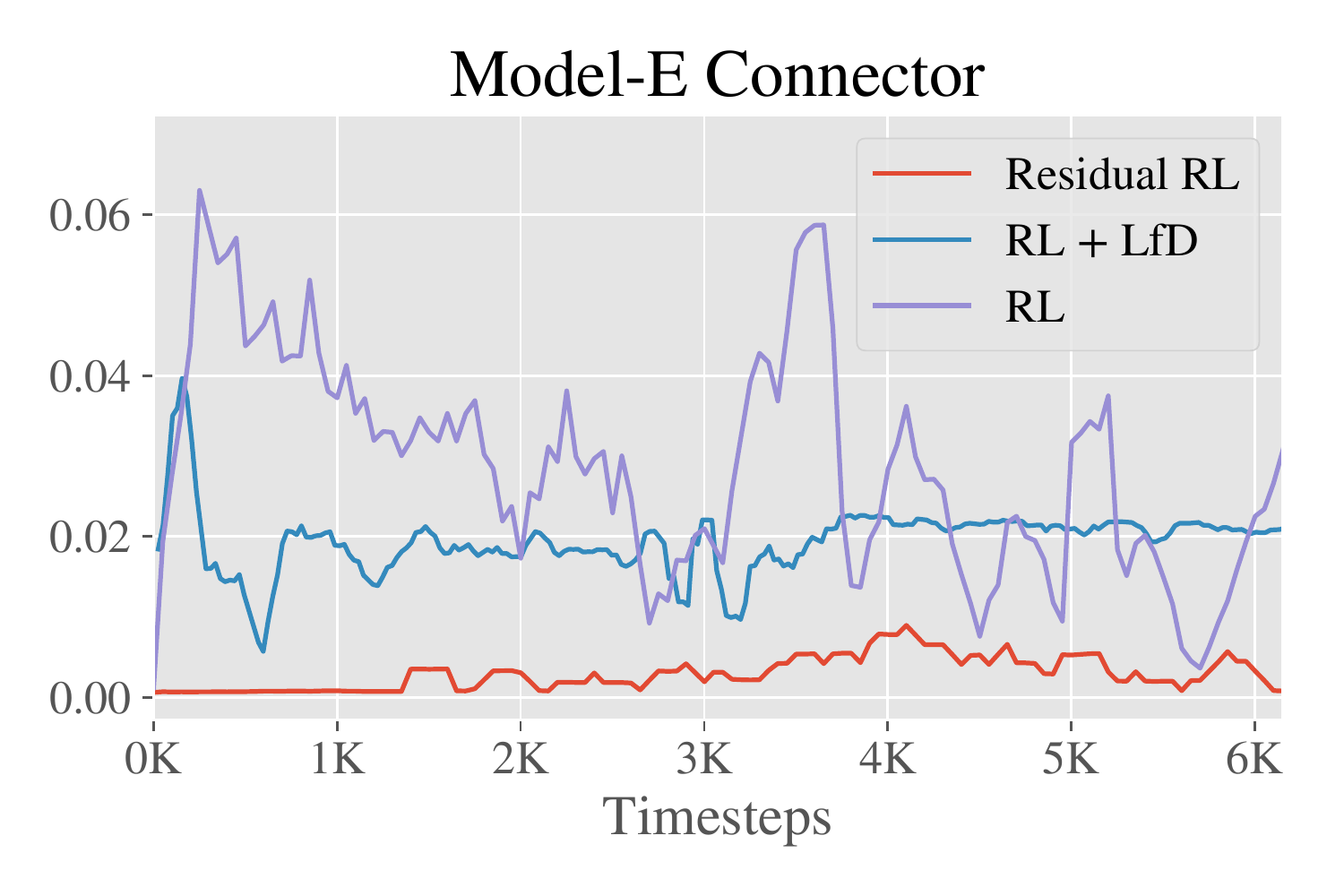}
    \end{subfigure}
    }
    \caption{Resulting final mean distance during the vision-based training. The comparison includes RL, residual RL, and RL with learning from demonstrations. Only residual RL manages to deal with the high-dimensional input and consistently solve all the tasks after the given amount of training. The other methods learn to move downwards, but often get stuck in the beginning of the insertion and fail to recover from unsuccessful attempts.}
    \label{vision_based_distance_all}
    \vspace{-0.5cm}
\end{figure}

\pagebreak

\subsection{Learning from Demonstrations}

Another method to incorporate prior information is to use demonstrations from an expert policy to guide exploration during RL. We first collected demonstrations with a joystick controller. Then, we add a behavior cloning loss while performing RL that pushes the policy towards the demonstrator actions, as previously considered in \cite{nair2018demonstrations}. Instead of DDPG, the underlying algorithm RL algorithm used is TD3.

\section{Experiments}\label{sec:experiments}

We evaluate our method, which combines residual RL with easy-to-obtain reward signals, on a variety of connector assembly tasks performed on a real robot. In this section, we consider two types of natural rewards that are intuitive for users to provide: an image directly specifying a goal and a binary sparse reward indicating success. For both cases, we report success rates on tasks they solve.  We aim to answer the following questions:  (1)~Can such trained policies provide comparable performance to policies that are trained with densely-shaped rewards? (2)~Are these trained policies robust to small variations and noise? 

\textbf{5.1 Vision-based Learning.} For the vision-based learning experiments, we use only raw image observations and $\ell_1$ distance between the current image and goal image as the reward signal. Sample images that the robot received are shown in Fig.~\ref{fig:vision_insertion_sequence}. We evaluate this type of reward on all three connectors. In our experiments, we use $32 \times 32$ grayscale images. 

\textbf{5.2 Learning from Sparse Rewards.} In the sparse reward experiment, we use the binary signal of the connector being electrically connected as the reward signal. This experiment is most applicable to electronic manufacturing settings where the electrical connection between connectors can be directly measured. We only evaluate the sparse reward setting on the USB connector, as it was straightforward to obtain the electrical connection signal.

\textbf{5.3 Perfect State Information.} After evaluating the tasks in the above settings, we further evaluate with full state information with a dense and carefully shaped reward signal, given in Eq.~\ref{eq:shaped_reward_function}, that incorporates distance to the goal and force information. Evaluating in this setting gives us an ``oracle'' that can be compared to the previous experiments in order to understand how much of a challenge sparse or image rewards pose for various algorithms.

\textbf{5.4 Robustness.} For safe and reliable future usage, it is required that the insertion controller is robust against small measurement or calibration errors that can occur when disassembling and reassembling a mechanical system. In this experiment, small goal perturbations are introduced in order to uncover the required setup precision of our algorithms.

\textbf{5.5 Exploration Comparison.} One advantage of using reinforcement learning is the exploratory behavior that allows the controller to adapt from new experiences unlike a deterministic control law.
The two RL algorithms we consider in this paper, SAC and TD3, explore differently. SAC maintains a stochastic policy, and the algorithm also adapts the stochasticity through training. TD3 has a deterministic policy, but uses another noise process (in our case Gaussian) to inject exploratory behavior during training time.
We compare the two algorithms, as well as when they are used in conjunction with residual RL, in order to evaluate the effect of the different exploration schemes.

\begin{figure}[btp]
\small
    \centering
        \begin{tabular}{|l|l|l|l|}
        \hline
        \multicolumn{2}{|l|}{\multirow{2}{*}{D-Sub Connector}} & \multicolumn{2}{c|}{Goal}                                 \\ \cline{3-4} 
        \multicolumn{2}{|l|}{}                     & \multicolumn{1}{c|}{Perfect} & \multicolumn{1}{c|}{Noisy} \\ \hline \hline
        \multirow{3}{*}{Pure RL} & Dense & 16\% & 0\% \\ 
         & Images, SAC & 0\% & 0\% \\
          & Images, TD3 & 12\% & 12\% \\ \hline
        \multirow{1}{*}{RL + LfD} & Images & 52\% & 52\% \\  \hline
        \multirow{3}{*}{Residual RL} & Dense & \textbf{100\%} & 60\% \\ 
         & Images, SAC & \textbf{100\%} & \textbf{64\%} \\
          & Images, TD3 & 52\% &  52\% \\ \hline
        \multirow{1}{*}{Human} & P-Controller & \textbf{100\%} & 44\% \\   \hline
        \end{tabular}
        \hfill
        \begin{tabular}{|l|l|l|l|}
        \hline
        \multicolumn{2}{|l|}{\multirow{2}{*}{Model-E Connector}} & \multicolumn{2}{c|}{Goal}                                 \\ \cline{3-4} 
        \multicolumn{2}{|l|}{}                     & \multicolumn{1}{c|}{Perfect} & \multicolumn{1}{c|}{Noisy} \\ \hline \hline
        \multirow{3}{*}{Pure RL} & Dense & 0\% & 0\% \\ 
         & Images, SAC & 0\% & 0\% \\ 
         & Images, TD3 & 0\% & 0\% \\ \hline
        \multirow{1}{*}{RL + LfD} & Images & 20\% & 20\% \\  \hline
        \multirow{3}{*}{Residual RL} & Dense & \textbf{100\%} & \textbf{76\%} \\ 
         & Images, SAC &\textbf{100\%} & \textbf{76\%} \\ 
           & Images, TD3 & 0\% & 0\% \\ \hline
        \multirow{1}{*}{Human} & P-Controller & 52\% & 24\% \\   \hline
        \end{tabular}
    \caption{We report average success out of 25 policy executions after training is finished for each method. For noisy goals, noise is added in form of $\pm 1\,\mathrm{mm}$ perturbations of the goal location. Residual RL, particularly with SAC, tends to be the best performing method across all three connectors. For the \text{Model-E}~connector, only residual RL solves the task in the given amount of training time.}
    \label{fig:tables}
\end{figure}

\section{Results}\label{sec:results}

We analyze the performance of policies learned with residual RL, as well as other methods, based on their ability to achieve the task goal, as well as the distance of the final object location to the goal pose over the course of training. To study the robustness of the learned policies, we also evaluate them in conditions where the goal connector position is perturbed, in order to understand the tolerance of RL policies to imprecise object placement.

\subsection{Vision-based Learning}
The results of the vision-based experiment are shown in Fig.~\ref{vision_based_distance_all}.
Our experiments show that a successful and consistent vision-based insertion policy can be learned from relatively few samples using residual RL. 

\begin{wrapfigure}{r}{0.5\linewidth}
    \label{tab:goal_pertubation_USB}
    \renewcommand{\arraystretch}{1.0}
    \begin{center}
    \small
    \begin{tabular}{|l|l|l|l|}
    \hline
    
    \multicolumn{2}{|l|}{\multirow{2}{*}{USB Connector}} & \multicolumn{2}{c|}{Goal}                                 \\ \cline{3-4} 
    \multicolumn{2}{|l|}{}                     & \multicolumn{1}{c|}{Perfect} & \multicolumn{1}{c|}{Noisy} \\ \hline \hline
    
    \multirow{5}{*}{Pure RL}   
     & Dense & 28\% & 20\% \\ 
     & Sparse, \;SAC & 16\% & 8\% \\ 
     & Sparse, \;TD3 & 44\% & 28\% \\ 
      & Images, SAC & 36\% & 32\% \\ 
       & Images, TD3 & 28\% & 28\% \\ \hline
    \multirow{2}{*}{RL + LfD} & Sparse &\textbf{100\%} & 32\% \\ 
     & Images & 88\% & 60\% \\ \hline
    \multirow{5}{*}{Residual RL} & Dense &\textbf{100\%} & \textbf{84\%} \\ 
     & Sparse,\: SAC & 88\% &  \textbf{84\%}\\ 
     & Sparse,\: TD3 &\textbf{100\%} & 36\% \\
     & Images, SAC & \textbf{100\%} & 80\% \\
      & Images, TD3 &  0\%& 0\%\\
      \hline
    \multirow{1}{*}{Human} & P-Controller & \textbf{100\%} & 60\% \\   \hline
    \end{tabular}
    \end{center}
    \vspace{5pt}
    \caption{Average success rate on the USB insertion task. Residual RL and RL + LfD solve the task consistently. Moreover, residual RL stays robust under $\pm1$\text{mm} noise. }
    \label{fig:TableUSB}
\end{wrapfigure}

This result suggests that goal-specification through images is a practical way to solve these types of industrial tasks. Although image-based rewards are often very sparse and hard to learn from, in this case the distance between images corresponds to a relatively dense reward signal which is sufficient to distinguish the different stages of the insertion process.

Interestingly, during training with standard RL, the policy would sometimes learn to ``hack'' the reward signal by moving down in the image in front of or behind the socket. In contrast, the stabilizing human-engineered controller in residual RL provides sufficient horizontal control to prevent this. The initial controller also scaffolds the learning process, by providing a very strong initialization that requires the reinforcement learning algorithm to only learn the final phase of the insertion. This produces substantially better performance in conjunction with vision-based rewards.

\begin{wrapfigure}{r}{0.5\linewidth}
    \centering
        \includegraphics[width=0.99\linewidth]{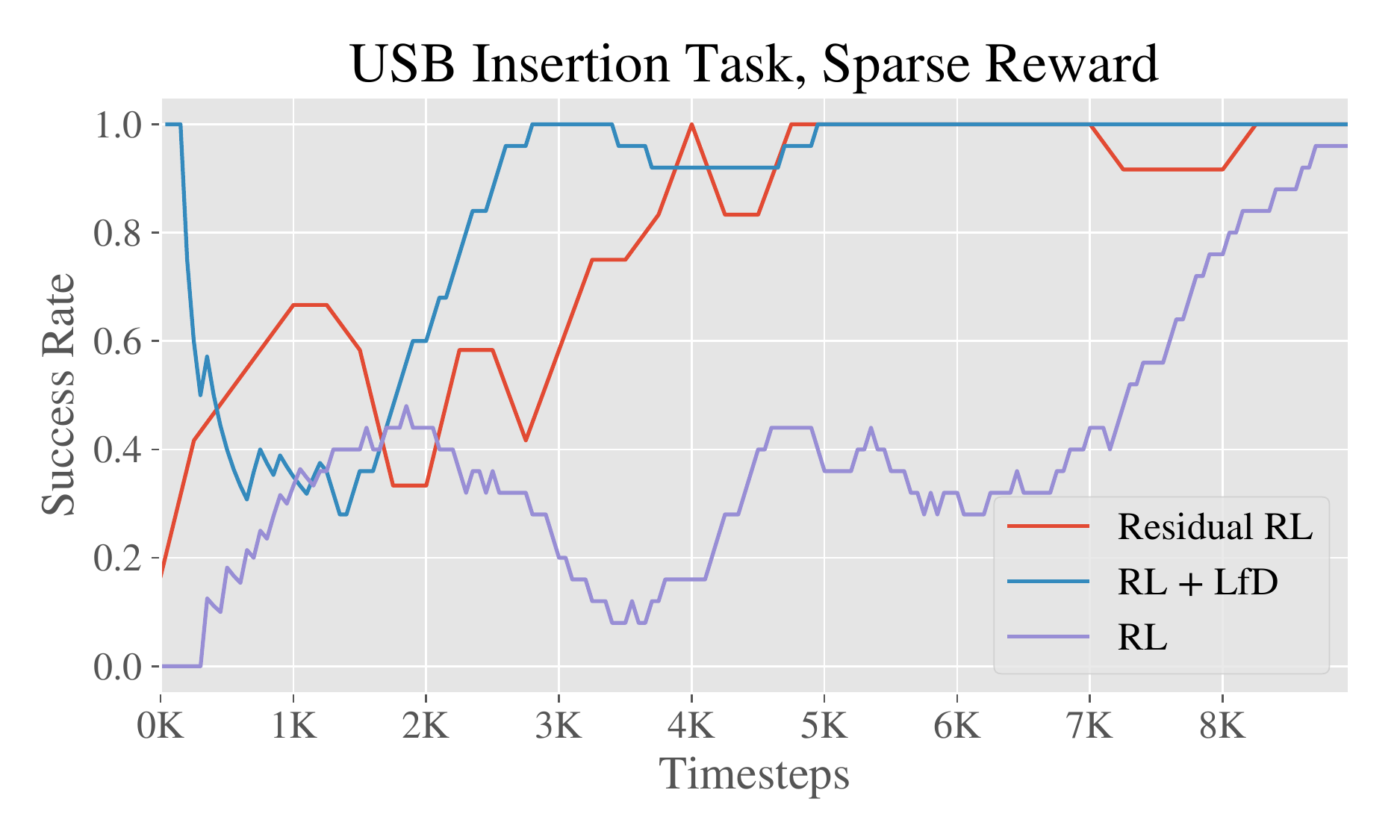}
    \caption{Learning curves for solving the USB insertion task with a sparse reward. Final distance to goal is shown; lower is better. Residual RL and RL with learning from demonstrations both solve the task relatively quickly, while RL alone takes about twice as long to solve the task at the same performance. }
    \label{fig:SparseRewardsAll}
    \vspace{-1.0cm}
\end{wrapfigure}

\subsection{Learning From Sparse Rewards}

In this experiment, we compare these methods on the USB insertion task with sparse rewards. The results are reported in Fig.~\ref{fig:SparseRewardsAll}. All methods are able to achieve very high success rates in the sparse setting. 
This result shows that we can learn precise industrial insertion tasks in sparse-reward settings, which can often be obtained much more easily than a dense, shaped reward. 
In fact, prior work has found that the final policy for sparse rewards can outperform the final policy for dense rewards as it does not suffer from a misspecified objective \cite{andrychowicz2017her}.

\subsection{Perfect State Information}
\begin{figure}[tbp]
    \centering
    \resizebox{1.02\linewidth}{!}{
    \begin{subfigure}[b]{0.32\linewidth}
        \includegraphics[width=0.99\linewidth]{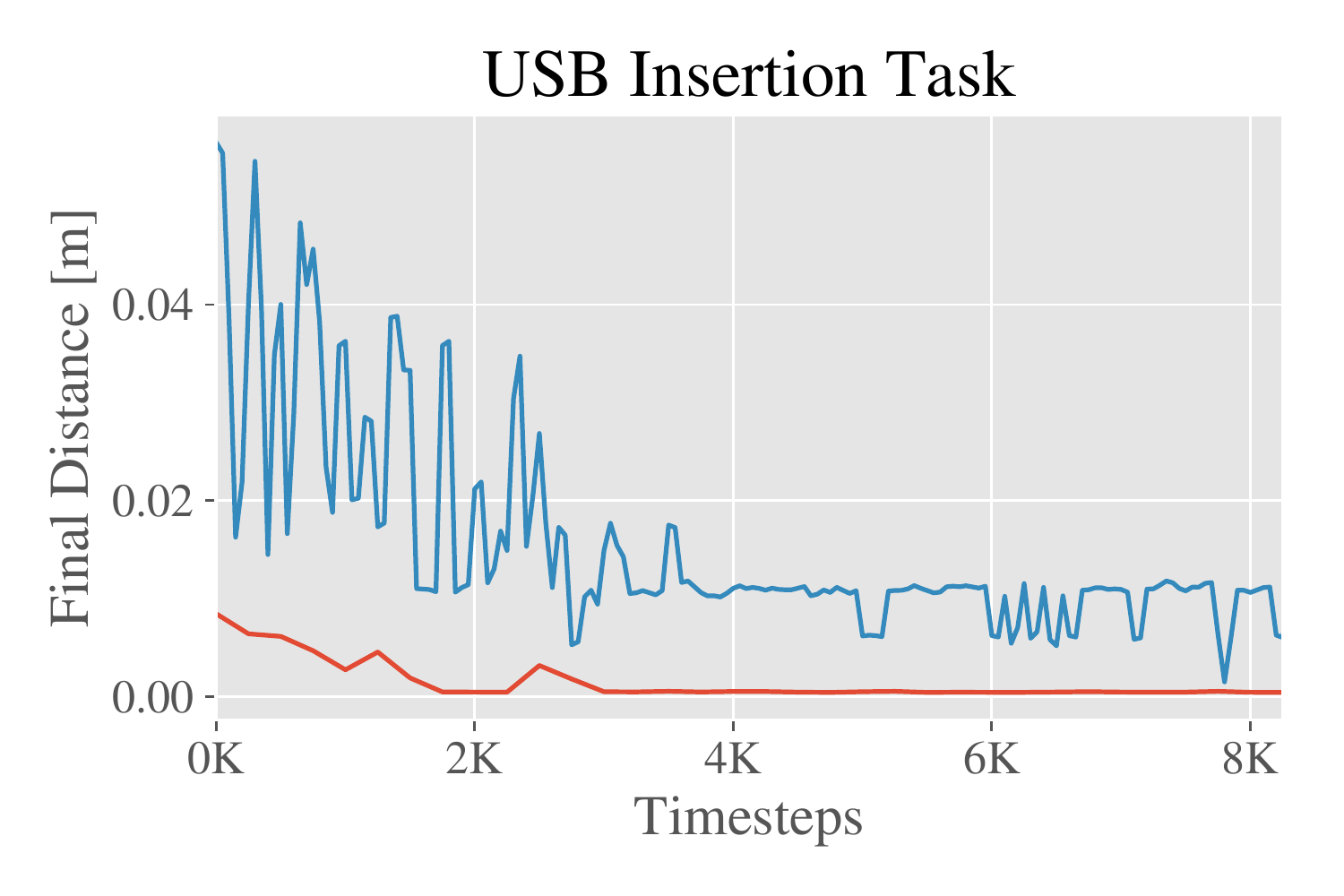}
    \end{subfigure}
    \begin{subfigure}[b]{0.32\linewidth}
        \includegraphics[width=0.99\linewidth]{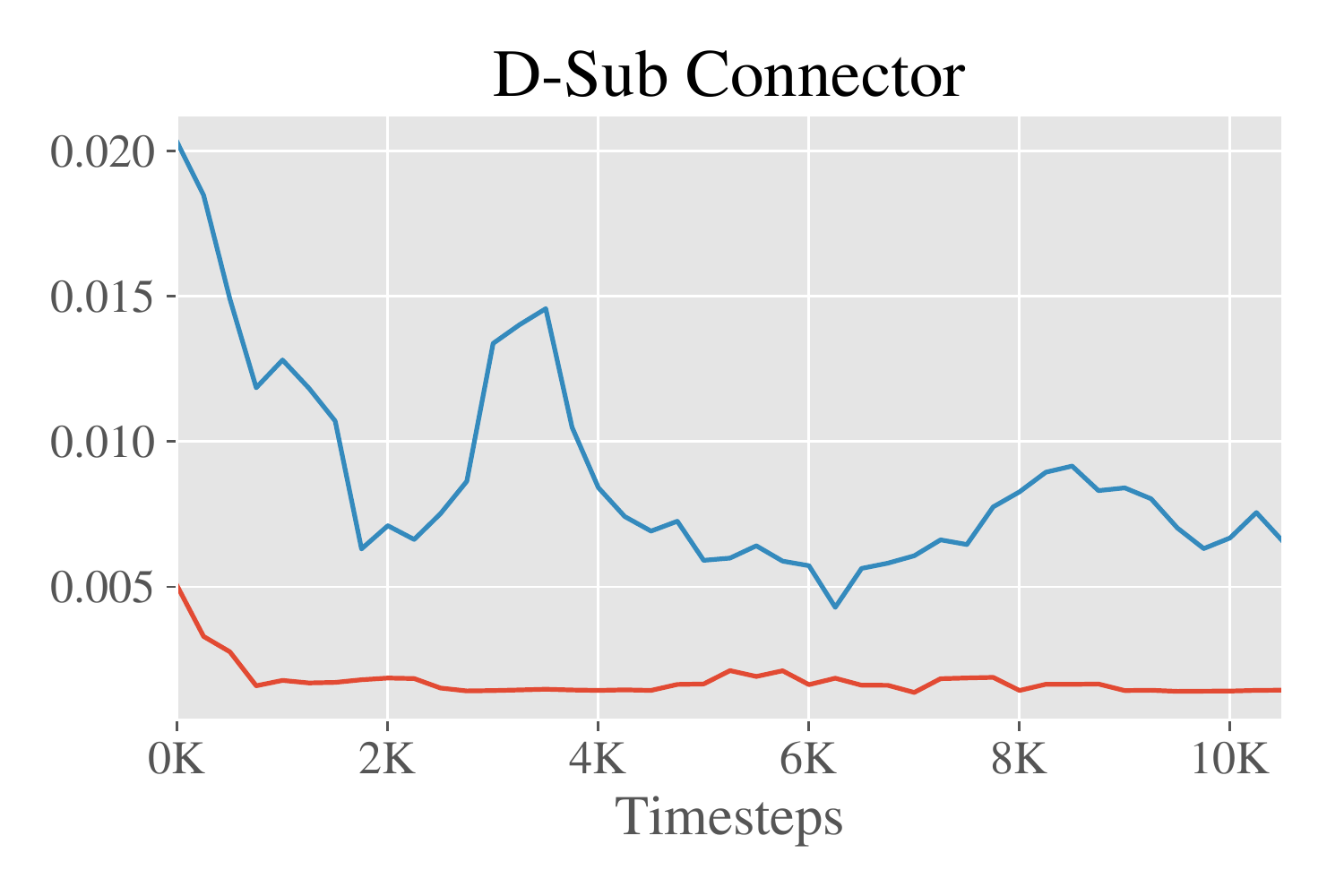}
    \end{subfigure}
    \begin{subfigure}[b]{0.32\linewidth}
        \includegraphics[width=0.99\linewidth]{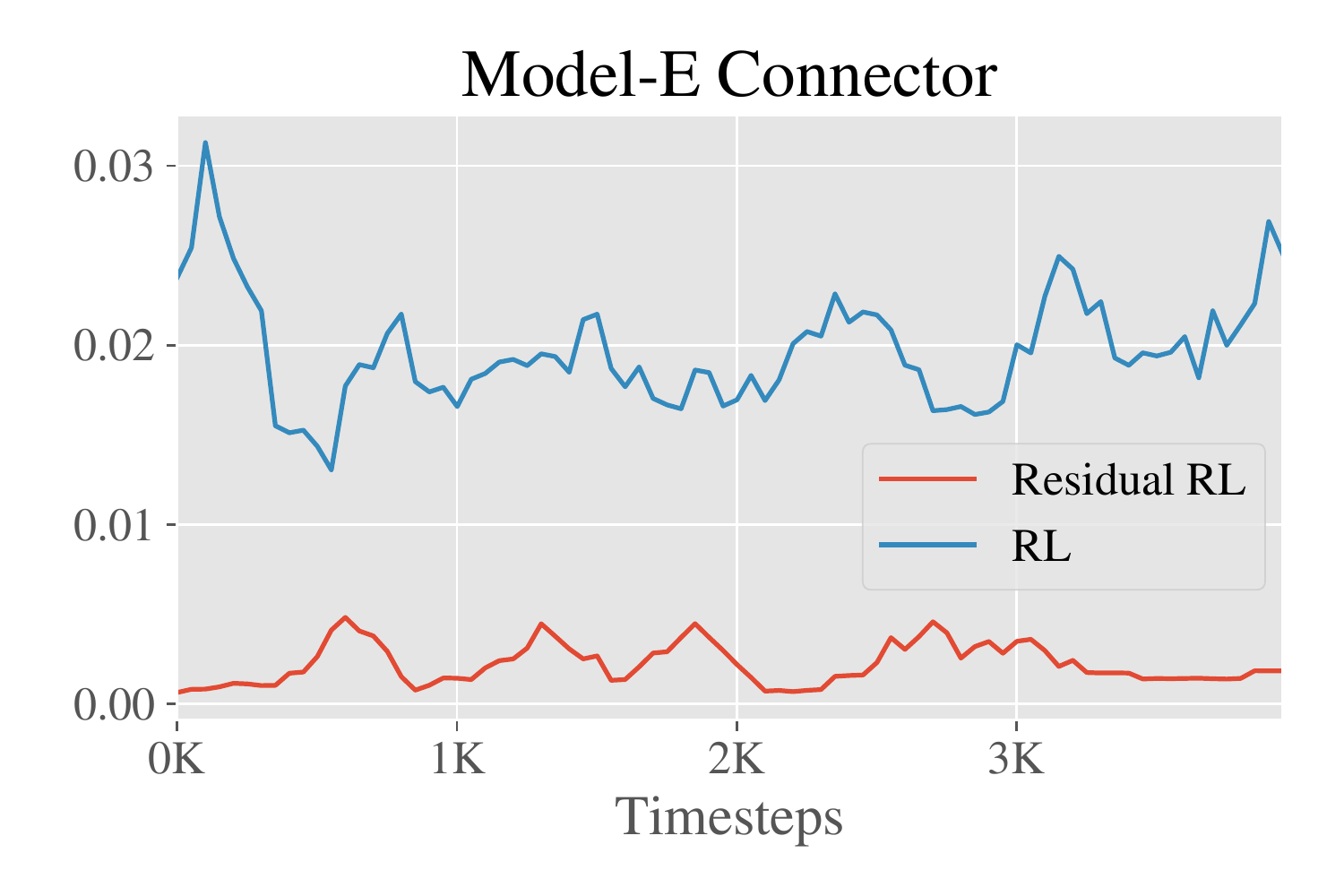}
    \end{subfigure}
    }
    \caption{Plots of the final mean distance to the goal during the state-based training. Final distances greater than $0.01\,\mathrm{m}$ indicate unsuccessful insertions. Here, the residual RL approach performs noticeably better than pure RL and is often able to solve the task during the exploration in the early stages of the training.}
    \label{fig:state_based_distance_all}
\end{figure}

The results of the experiment with perfect state information and dense rewards is shown in Fig.~\ref{fig:state_based_distance_all}. In this case, residual RL still outperforms standard RL, though the better-shaped reward enables standard RL to make more initial progress than with the other reward signals.
However, the hand-designed shaped reward function makes it harder for the policy to actually perform the full insertion, potentially because the more complex reward landscape provides other competing goals to the policy. The final performance with sparse rewards on the USB insertion task is substantially better.

\subsection{Robustness}

In the previous set of experiments, the goal locations were known exactly.
In this case, the hand-engineered controller
performs well. However, once
noise is added to the goal location, the deterministic P-controller struggles. 
To test robustness, a goal perturbation is created artificially, and the controllers are tested under this condition. 
All results of our robustness evaluations are listed in Fig.~\ref{fig:tables} and Fig.~\ref{fig:TableUSB}.  
In the presence of a $\pm 1\mathrm{mm}$ perturbation, the
residual RL 
controller succeeds more often on the USB and D-Sub tasks, and significantly more often on the Model-E task.
Unlike the hand-engineered controller, residual RL consistently solved this task and overcame goal perturbations in $16/25$ trials. 
The agent demonstrably learns small but consistent corrective
feedback behaviors in order to move in
the right direction during the descent motion, a behavior that
is very difficult to specify manually.
This behavior illustrates the strength of
residual RL. Since the human controller already specifies the general
trajectory of the optimal policy, environment samples are only
required to learn this corrective feedback behavior.

\subsection{Exploration Comparison}
All experiments were also performed using TD3 instead of SAC. 
The final success rates of these experiments are included in Fig.~\ref{fig:tables}. 
When combined with residual RL, SAC and TD3 perform comparably. 
However, TD3 is often substantially less robust.
These results are likely explained by the exploration strategy of the two algorithms. 
TD3 has a deterministic policy and fixed noise during training, so once it observes some high-reward states, it quickly learns to repeat that trajectory. 
SAC adapts the noise to the correct scale, helping SAC stay robust to small perturbations, and because SAC learns the value function for a stochastic policy, it is able to handle some degree of additive noise effectively.
We found that the outputted action of TD3 approaches the extreme values at the edge of the allowed action space, while SAC executed less extreme actions, which likely further improved robustness.

\section{Conclusion}\label{sec:discussion}
In this paper, we studied deep reinforcement learning in a practical setting, and demonstrated that deep RL can solve complex industrial assembly tasks with tight tolerances.
We showed that we can learn insertion policies with raw image observations with either binary outcome-based rewards, or rewards based on on goal images.
We conducted a series of experiments for various connector type assemblies, and demonstrated the feasibility of our method under challenging conditions, such as noisy goal specification and complex connector geometries.
Reinforcement learning algorithms that can automatically learn complex assembly tasks with easy-to-specify reward functions have the potential to automate a wide range of assembly tasks, making this technology a promising direction forward for flexible and capable robotic manipulators.

There remains significant challenges for applying these techniques in more complex environments. One practical direction for future work is focusing on multi-stage assembly tasks through vision. This would pose a challenge to the goal-based policies as the background would be visually more complex. Moreover, multi-step tasks involve adapting to previous mistakes or inaccuracies, which could be difficult but should be able to be handled by RL. 
Extending the presented approach to multi-stage assembly tasks will pave the road to a higher robot autonomy in flexible manufacturing.

\section{Acknowledgements}
This work was supported by the Siemens Corporation,
the Office of Naval Research under a Young Investigator
Program Award, and Berkeley DeepDrive.


{ \small
\bibliographystyle{IEEEtran}
\bibliography{example}
}

\end{document}